\documentclass[conference]{IEEEtran}
\IEEEoverridecommandlockouts
\usepackage{cite}
\usepackage{amsmath,amssymb,amsfonts}
\usepackage{algorithmic}
\usepackage{graphicx}
\usepackage{textcomp}
\usepackage{xcolor}
\usepackage{subfigure}
\usepackage{bm}
\usepackage{dblfloatfix} 
\def\BibTeX{{\rm B\kern-.05em{\sc i\kern-.025em b}\kern-.08em
    T\kern-.1667em\lower.7ex\hbox{E}\kern-.125emX}}

\begin{document}

\title{Modeling human road crossing decisions as reward maximization with visual perception limitations }
%\centering{
\author{Yueyang Wang$^{1\star}$,~Aravinda Ramakrishnan Srinivasan$^{1}$,~Jussi P.P. Jokinen$^{2}$,~Antti Oulasvirta$^{3}$,~Gustav Markkula$^1$\\
\thanks{$^{*}$ Corresponding author: {\tt\small mn20yw2@leeds.ac.uk}}%
\thanks{This project has received funding from UK Engineering and Physical Sciences Research Council under fellowship  COMMOTIONS - Computational Models of Traffic Interactions for Testing of Automated Vehicles - EP/S005056/1.}
{$^{1}$~University of Leeds, UK 
$^{2}$~University of Jyv\"{a}skyl\"{a}, Finland
      $^{3}$~Aalto University, Finland}}

% \and
% \and
% \and
% \and
% \and

\maketitle

\begin{abstract}
Understanding the interaction between different road users is critical for road safety and automated vehicles (AVs). Existing mathematical models on this topic have been proposed based mostly on either cognitive or machine learning (ML) approaches. However, current cognitive models are incapable of simulating road user trajectories in general scenarios, and ML models lack a focus on the mechanisms generating the behavior and take a high-level perspective which can cause failures to capture important human-like behaviors. Here, we develop a model of human pedestrian crossing decisions based on computational rationality, an approach using deep reinforcement learning (RL) to learn boundedly optimal behavior policies given human constraints, in our case a model of the limited human visual system. We show that the proposed combined cognitive-RL model captures human-like patterns of gap acceptance and crossing initiation time. Interestingly, our model’s decisions are sensitive to not only the time gap, but also the speed of the approaching vehicle, something which has been described as a “bias” in human gap acceptance behavior. However, our results suggest that this is instead a rational adaption to human perceptual limitations. Moreover, we demonstrate an approach to accounting for individual differences in computational rationality models, by conditioning the RL policy on the parameters of the human constraints. Our results demonstrate the feasibility of generating more human-like road user behavior by combining RL with cognitive models.
\end{abstract}


\section{Introduction}
The interaction between road users is defined as ``a situation where the behavior of at least two road users can be interpreted as being influenced by the possibility that they are both intending to occupy the same region of space at the same time in the near future''~\cite{1markkula2020defining}. Interdependence between vehicles and pedestrians makes interactions between road users instrumental for road safety and automated vehicles(AVs), which pushes the research into road user interaction. 

Pedestrian is the most vulnerable group among all road users~\cite{2reported2021}, and their behavior is difficult to predict. How drivers may behave is limited by the machinery of a vehicle and traffic rules, whereas pedestrians have more freedom, and are limited only by traffic rules. To better understand pedestrian behavior, some detailed metrics related to crossing behavior were investigated~\cite{5tian2022explaining,8lobjois2007age,3petzoldt2014relationship}. For example, gap acceptance, where the gap is defined as the time or spatial distance between the pedestrian and approaching vehicle, is an important metric for understanding the crossing decision~\cite{5tian2022explaining}. Lobjois and Cavallo~\cite{8lobjois2007age} found that speed-dependent gap acceptance was shown in different age groups i.e., the gap acceptance rate was higher when the approaching vehicle was faster in a given time gap. Petzoldt~\cite{3petzoldt2014relationship} investigated the relationship between gap acceptance and time to arrival (TTA) estimation. They speculated that speed-dependent crossing decisions were caused by the biased TTA estimation. Another important factor affecting the safety of crossing is cross initiation time (CIT). Tian et al.~\cite{5tian2022explaining} observed that CIT was greater when the vehicle was driven at a higher speed for any given initial TTA, which led to unsafe behavior. 

A descriptive study of road user behavior is insufficient for AVs to understand and predict other road users’ actions; therefore, mathematical models of road user behavior are required for AVs~\cite{9camara2020pedestrian}. In recent years, many mechanistic models have been proposed to generate and understand pedestrian behavior. For example, rule-based models, such as the social force model, were successful in traffic flow simulation~\cite{10helbing1995social,11yang2020social}. However, they are limited in capturing the details of road user interactions. To generate more explainable and accurate road user interactive behavior, cognitive models, such as the evidence accumulation model, were utilized to model pedestrian crossing decisions~\cite{12markkula2018models,13giles2019zebra}. 

With increased computing power and more available data, machine learning (ML) models have gained increasing attention for road user behavior prediction. Long Short-Term Memory (LSTM), a variant of Recurrent Neural Networks (RNN), was used in pedestrian trajectory prediction~\cite{16wang2018eidetic}. To better predict the interactive behavior between pedestrians, Alahi et al proposed a ‘Social LSTM’, with a social layer into the LSTM algorithm, and the model outperformed state-of-the-art methods~\cite{18alahi2016social}. 

Many efforts have been made to understand and simulate the microscopic behavior of road users. ML models can reproduce accurate trajectories across a diverse range of scenarios, and cognitive models can provide the interpretability of the interactive behavior and the underlying mechanism. However, both streams of methods have some limitations. In conventional cognitive models, the modeler should define how the specific task is completed and the model should be updated if the environment and task change. This makes it difficult to simulate road users’ trajectories in general scenarios. Whereas ML models focus on the minimization of high-level error metrics, rather than the mechanisms generating the behavior or whether the aspects of behavior that are important to humans are being captured. Sometimes the model with high accuracy does not necessarily generate realistic human behavior~\cite{20srinivasan2022beyond}.  

Computational rationality, as a general approach of modeling human behavior, has shown promising properties in modeling human-computer interaction (HCI)~\cite{21oulasvirta2022computational}. This framework is based on the idea that human behaviors are generated by cognitive mechanisms that are adapted to the structure both of the environment and the mind and brain itself~\cite{22lewis2014computational}. In this paper, we developed a model of human pedestrian crossing decisions based on computational rationality, using deep reinforcement learning (RL) to adopt optimal behavior policies given human-like constraints. We show that when we constrain the agent by a simple model of human visual perception, it reproduces human gap acceptance behavior qualitatively, including the speed-dependencies, which have not been previously considered as rational behavior. We also demonstrate an approach to using computational rationality to model individual differences, by conditioning RL on the parameters of human constraints. 
\section{Method}
\subsection{Dataset}

\begin{figure}[]
      \centering
      \includegraphics[scale=0.4]{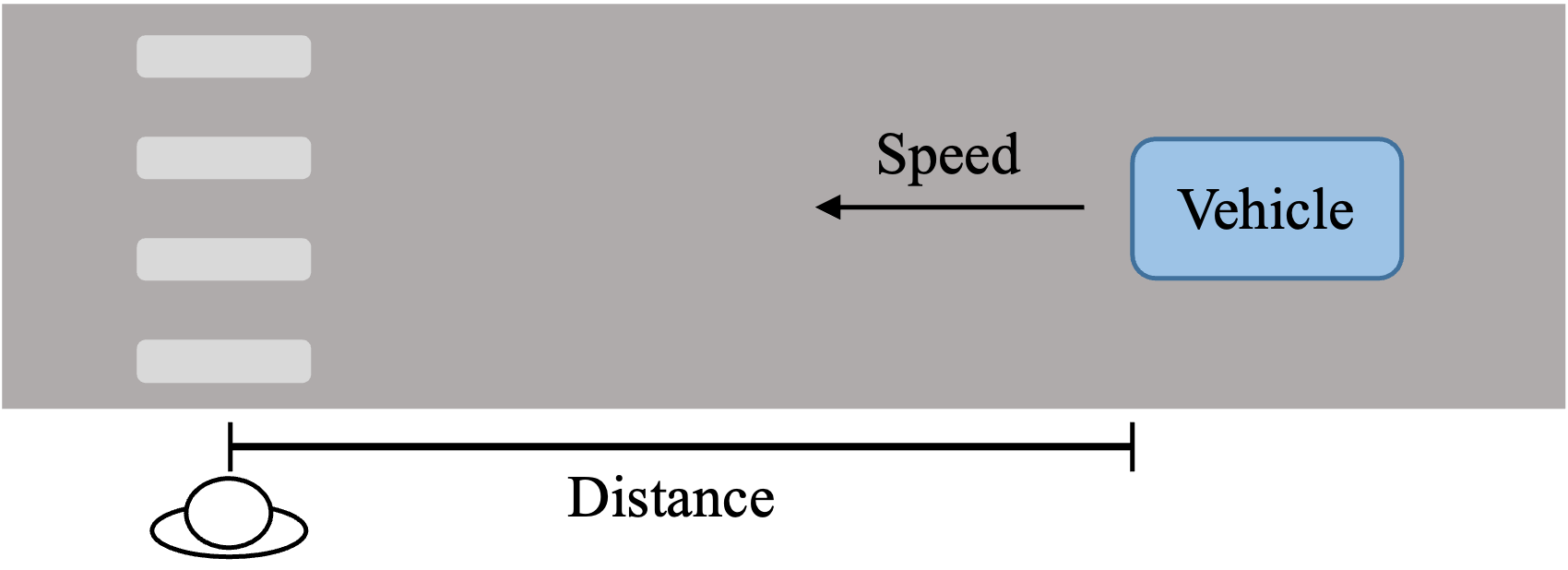}
      \vspace{-0.2cm}
      \caption{Birds-eye view of the experiment.}
      \label{fig:birdview}
      \vspace{-0.4cm}
\end{figure}

The dataset used for validation was collected in the previous experiment reported by Giles et al.~\cite{13giles2019zebra}. \figurename~\ref{fig:birdview} shows a birds-eye view of the experiment. A brief summary of the experimental setup is provided below. 20 participants were recruited for the experiment. In the experiment, they wore an HTC Vive Virtual Reality (VR) headset and experienced the virtual crossing task. The VR environment consisted of a straight two-lane road with a total width of $5.85$~m, with a zebra crossing at the participant’s location. 

In terms of the experimental procedure, participants stood in front of the zebra crossing. When participants were ready to start the trial, they turned their head to the right to trigger the scene. A car at the predefined initial position $d_0$ would approach the participant at different speeds $v_0$. The experiment included a mix of scenarios; in this paper, we will only consider the scenarios where the speed of the approaching vehicle was constant. The detail of these scenarios is shown in~\tablename~\ref{tab:my_label}, where also the initial time to arrival (TTA) $\tau_0 = d_0/v_0$ is listed. Participants pressed the button on the HTC Vive’s controller when they felt safe to cross. Upon this button press, CIT was recorded, and the location of the participant in the virtual environment moved across the zebra crossing at the speed of $1.31$~m/s. This button-press approach was chosen in favor of physically crossing the road, to reduce the impact of variability in motor constraints on the crossing decision. Each participant experienced $6$ different constant-speed trials. Therefore, $120$ data trials were used for the validation of the model.

\subsection{Model}
\begin{figure}[t]
    % \vspace{-0.2cm}
      \centering
      \subfigure[]{
      \includegraphics[scale=0.4]{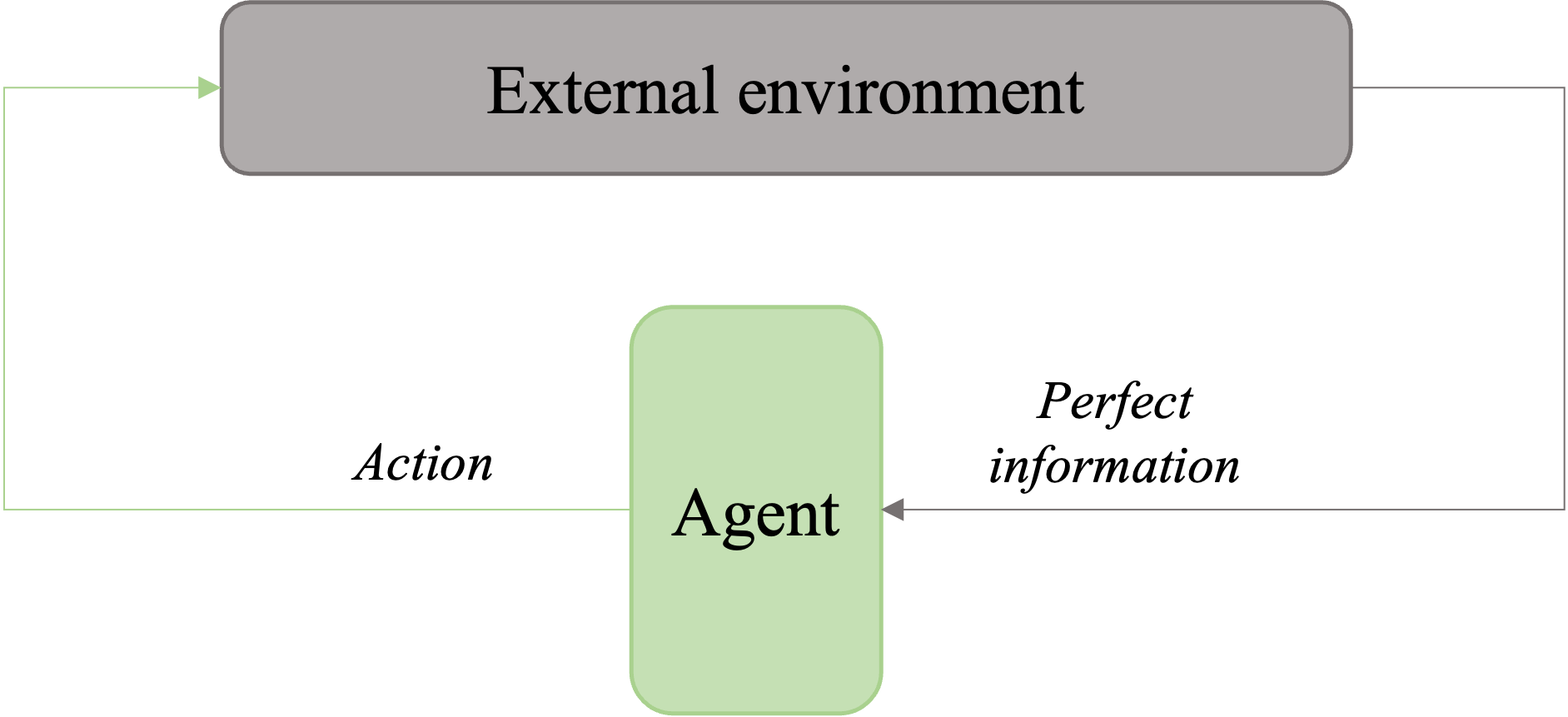}\label{fig:idealmod}}
      %\centering
      \subfigure[]{
      \includegraphics[scale=0.4]{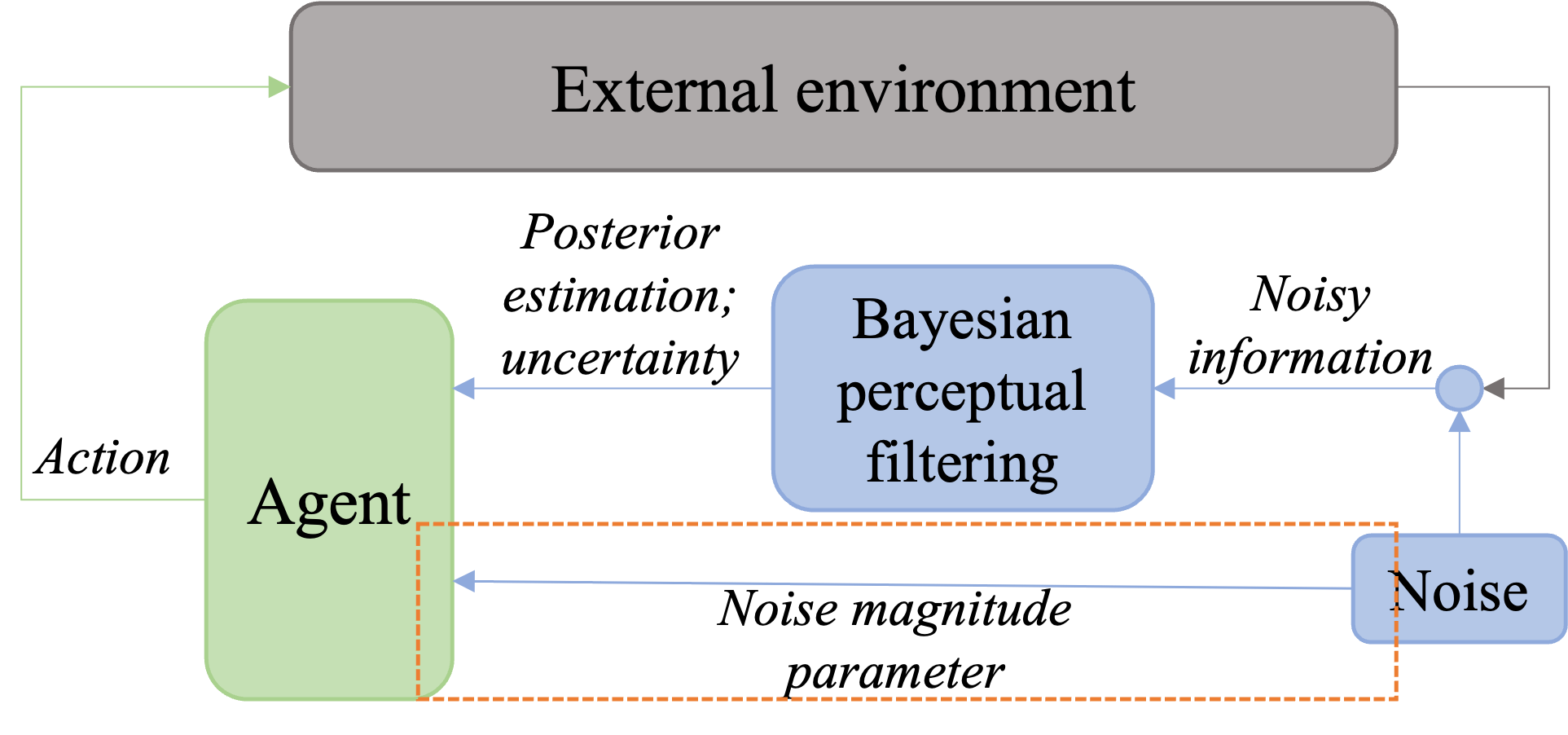}\label{fig:crmodel}}
      \vspace{-0.2cm}
      \caption{Comparison of models. (a) Ideal observer model, where the agent had full information about the environment. (b) Model with visual limitations. Two variants were developed. One was the model without noise magnitude parameter. Another was the model with noise magnitude parameter, as shown in the orange dashed box.}
      \label{fig:models}
      \vspace{-0.2cm}
\end{figure}

This research aims to model the pedestrian crossing decision under the assumption that human behaves rationally within limits. Therefore, two models were compared, as shown in~\figurename~\ref{fig:models}. One is the ideal observer crossing model, in which the agent has perfect information about the environment. Another is the model considering the visual limits. In this model, the agent perceives the environment subject to noise, but Bayes-optimal perception. 
\begin{table}[b]
    \vspace{-0.2cm}
    \caption{Vehicle approach scenarios}
    \vspace{-0.2cm}
    \centering
    \begin{tabular}{|c|c|c|}
         \hline
        \bm{$v_0~m/s$}&\bm{$d_0~(m)$}&\bm{$\tau_0~(s)$} \\ [0.2ex]
         %\specialcell{$v_0~m/s$}&\specialcell{$d_0~(m)$}&\specialcell{$\tau_0~(s)$}  \\
         \hline
         $6.94$&$15.90$&$2.29$ \\
         \hline
         $13.89$&$31.81$&$2.29$ \\
         \hline
         $6.94$&$31.81$&$4.58$ \\
         \hline
         $13.89$&$63.61$&$4.58$ \\
         \hline
         $6.94$&$47.71$&$6.87$ \\
         \hline
         $13.89$&$95.42$&$6.87$ \\
         \hline
    \end{tabular}
    %\vspace{-0.4cm}
    \label{tab:my_label}
\end{table}
\subsubsection{Noisy perception}
We assume the agent has a noisy perception of the state of vehicles and perfect knowledge about their own state.
\paragraph{Noisy visual input}The observation obtained by the agent is according to the principle of the human visual system, i.e., the sensory input received by our human visual system is noisy~\cite{23faisal2008noise}. It is important to consider the nature of this noise; here we are building on models which assume that visual noise is introduced at the level of the human retina, as angular noise~\cite{24kwon2015unifying}. In the current model, we assume that the agent observes the position of the other agent along its line of travel by observing the angle below the horizon of the other agent~\cite{25ooi2001distance,26markkula2022explaining}, with a constant Gaussian noise of standard deviation $\sigma_v$. In practice, this means that the pedestrian observes the position of the vehicle with a distance-dependent noise of standard deviation $\sigma_x(k) = f_v[x(k)]$, where 
$x(k)$ is the true world state, and: 
\begin{equation*}
f_v[x(k)] = |d_l| \left( 1-\frac{h}{d\cdot\tan(\arctan\frac{h}{d}+\sigma_v)}\right),
\end{equation*}
where $d_l$ is the longitudinal distance between the pedestrian agent and the crossing point, $d$ is the distance between the agent and the approaching vehicle, $h$ is the eye height over the ground of the ego agent, which is set to $1.6$~m for all pedestrian agents for simplicity, and $\sigma_v$ could vary between pedestrians.

\paragraph{Kalman filter}
There is psychophysical evidence that human perception system works like a Bayesian optimizer, and Bayesian methods have been successful in modeling perception and sensorimotor control~\cite{24kwon2015unifying,27knill2004bayesian}. Therefore, we used a Kalman Filter as a model of the human visual perception to percept the environment~\cite{26markkula2022explaining}. In our model, we initialized the Kalman filter with a noisy position of the vehicle, and a noisy velocity centered at the true velocity with a standard deviation of all velocity values. At each step, the Kalman filter received the noisy position about the other agent, and the output, i.e., the filtered position and velocity of the vehicle, and the variance of the position and the velocity, was the input of the RL agent. 
\subsubsection{Reinforcement learning model}
In our model, in line with the theory of computational rationality, we view pedestrian behavior as a Partially Observable Markov Decision Process (POMDP) under bounds posed by perception. RL algorithm, where the agent interacts with the environment and learns the optimal strategy by trial and error, can be used to derive the boundedly optimal policy for this type of problem~\cite{21oulasvirta2022computational,28franccois2018introduction}. 
\paragraph{State space $S$}
The time step in the simulation affects the resolution of the results of the decision time. In our mode, one time step corresponds to $0.1$ seconds, which is suitable for the dataset we are using. At each time step t, the environment is in a state $s_t \in S$. A state contains true information about the vehicle and the agent, i.e., the position and velocity of the vehicle and the agent. 
\paragraph{Action space $A$}
At each time step t, the agent takes an action $a_t \in A$. In this paper, in line with the button press in the experiment, the agent can make the decision to Go or Not Go. If the Go decision is made, the agent will go straight at the speed of $1.31~m/s$, as in the experiment, resulting either in a successful crossing or in a collision, and the scenario will finish. 
\paragraph{Reward $R$}
In the experiment where the datasets were collected, the participant’s task was to cross the road as soon as they felt safe to do so, either before or after the car had passed them~\cite{29pekkanen2022variable}. Therefore, in our model we want the agent to cross the road in as short a time as possible without a collision. At each time step t, the agent will receive a negative reward of $0.5 \times simulation steps$, which helps the agent to cross the road faster. The agent will be given a reward of $200$ when crossing the road without collision, and a reward of $-200$ if a collision happens. The form of this reward function was chosen based on some initial testing. As the focus of this study is on the potential effect of noisy perception on the crossing decision, we kept the reward function simple; future work can further refine it to better capture human preferences. 

\paragraph{Observation space $O$}
The agent receives observation $o_t \in O $ at each time step. In the ideal observer model, the agent gets the complete information about the environment. In the model with visual limits, the agent only observes partial information about the state of the environment. At each time step, the agent receives the processed estimates of the position and velocity and uncertainty about the position and velocity of the other agent from the Kalman Filter and the exact position and velocity of the ego agent.

\begin{figure*}[b!]
      \vspace{-0.2cm}
      \centering
      \includegraphics[scale=0.45]{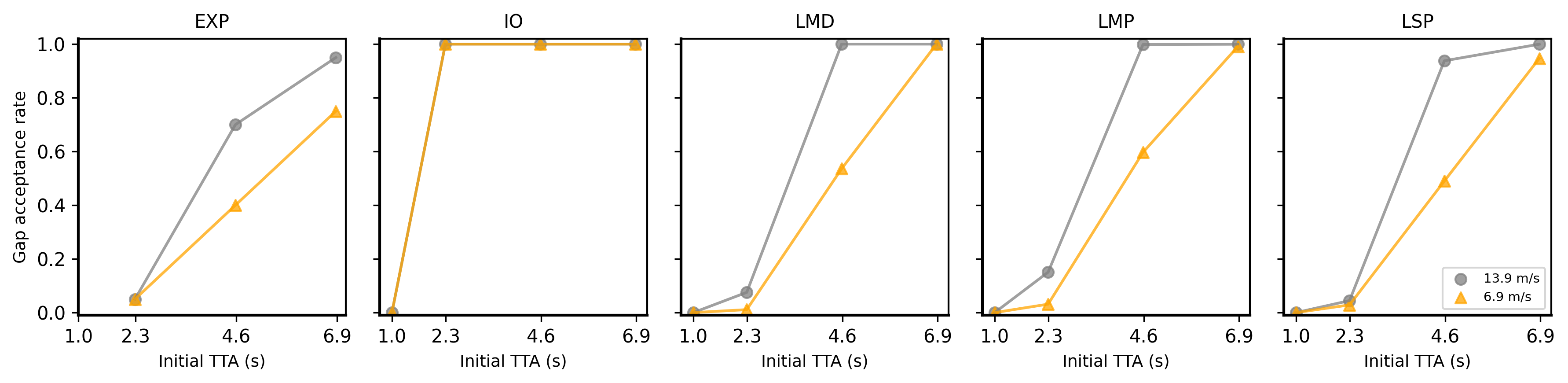}
      \vspace{-0.2cm}
      \caption{Gap acceptance rate by human participants and different models. See the main text for explanations of the model abbreviations.}
      % \vspace{-0.4cm}
      \label{fig:modelgap}
\end{figure*}

\paragraph{Transition function $T$}
The transition function defines how the current state $s_t$ changes to the next state $s_{t+1}$ taking action $a_t$. In our model, if Not Go action is chosen, the vehicle will move according to the kinematic equations with the given speed, and the position of the agent will not change. Once Go action is chosen, whether the collision happens will be calculated, and the corresponding reward will be given to the agent. Then, the simulation finishes. 
\paragraph{Deep Q-Networks}
Deep-Q Network (DQN) is a method using the neural network to learn the optimal policy to maximize the state-action function (Q function), $Q(s,a)$, the expected rewards for an action taken in a given state~\cite{31van2016deep}. DQN is suitable for the problem with a continuous state space and a discrete action space. For the extensibility of the model to more complex situations, we utilized an enhanced version of DQN, Double DQN (DDQN). The Double DQN (DDQN) structure, which decouples the update of the neural network parameter for action selection and evaluation, can avoid the overestimation of the action value~\cite{31van2016deep}. Furthermore, the dueling network was used, in which the Q-function is decoupled to a value function and a state-dependent action advantage $A(s,a)$ function. Compared with the single-stream DQN, the Duelling DQN shows better performance especially when different actions lead to a similar value because of the consideration of the state value in the Q value~\cite{32wang2016dueling}. We trained the agent through a two-layer fully connected network, with $512$ and $256$ nodes. The learning rate and discount factor are $0.001$ and $0.99$ respectively. To explore the optimal policy, an $\epsilon$ - greedy algorithm was used for exploration: At each time step $t$, a random action is chosen with probability $\epsilon$, and the action with maximum Q value is chosen with probability $1-\epsilon$. We decreased $\epsilon$ by $10^{-4}$ in each learning step. The minimum value was set to $0.001$. 

\subsection{Training and fitting}
As we don’t know the correct value for $\sigma_v$, and additionally each participant in the experiment may have an individual $\sigma_v$, we trained the model for different $\sigma_v$ ranging from $0-1$ with a step size of $0.002$. We trained the model in two different ways. First, we trained a separate model for each $\sigma_v$. With this approach, for any given $\sigma_v$, we can get a boundedly optimal policy from the model trained with that $\sigma_v$. This approach is somewhat inconvenient and will not scale well to more complete models of human constraints, with more parameters than just one. Therefore, as an alternative approach, we instead trained just a single model, across all of the different $\sigma_v$, and in this case we also provided $\sigma_v$ as an input to the model, as illustrated by the dashed line in~\figurename~\ref{fig:models}. In other words, we conditioned the RL on $\sigma_v$. With this approach, for any given $\sigma_v$, we get a boundedly optimal policy from this single model, by also giving it $\sigma_v$ as an input. 

With the $1.31~m/s$ walking speed used in the experiment, it is in fact possible to cross before the vehicle without collision even for the lowest TTA of $2.29~s$. To avoid our agents learning this trivial policy of always just immediately crossing, in the model training we also included a lower TTA of $1~s$, in which the agent did not have enough time to cross safely before the vehicle.  

We used three criteria to identify model convergence and stop training:(1) the collision rate is less than or equal to $0.01$, which means one collision at most happened in the last $100$ episodes; (2) $\epsilon$ has reached the lowest value of the epsilon-greedy algorithm; (3) the difference in average reward between the last $100$ episodes and the last $200$ to $100$ episodes is less than $1$ to make sure that the average reward does not improve much and becomes stable and converged. 

We also identified the $\sigma_v$ that fitted the experimental data best for each participant by likelihood maximization. We estimated the probability density function (PDF) of CIT predicted for each model by kernel density estimation, separately for each of the six scenarios (Table 1). This allowed us to calculate the model likelihood of each $\sigma_v$ for each participant, by multiplying the model PDF values at the participant’s observed button press times. Finally, we combined all the PDFs from the different participants, with tuned $\sigma_v$. We applied this method to both model types, i.e., separate models for each $\sigma_v$, and a single model for all $\sigma_v$. In addition to this per-participant fit, we also selected one $\sigma_v$ from the models with separate $\sigma_v$, which best fitted the entire dataset. 

Abbreviations will be introduced here for referring to different model variants; e.g., EXP: experimental data; IO: ideal observer model; LMD: multiples model with separate visual limits parameter and one $\sigma_v$ fitted across the entire dataset; LMP: multiple models with separate visual limits parameter and different $\sigma_v$ fitted per participant; LSP: one model with different visual limits parameters and different $\sigma_v$ fitted per participant. 
\section{Results}
\subsection{Experimental results}
First, we reanalyzed the experimental data reported by Pekkanen et al.~\cite{29pekkanen2022variable}. As shown in the first panel on the left in~\figurename~\ref{fig:modelgap}, this experiment replicated the general finding that the gap acceptance rate strongly depended on initial TTA, i.e., more pedestrians crossed before the car when there was a larger time gap. In addition, we can observe the speed-dependent gap acceptance rate, i.e., for a given initial TTA, more pedestrians accepted the gap if the speed was higher~\cite{5tian2022explaining,8lobjois2007age}. This effect was strongest at the initial TTA of $4.6~s$. Regarding CIT, which is shown in the first row in~\figurename~\ref{fig:modelcit}, there was some spread in CIT, both when crossing before and after the car. 

\subsection{Behavior of the model}
\begin{figure*}[!t]
      \centering
      \includegraphics[scale=0.4]{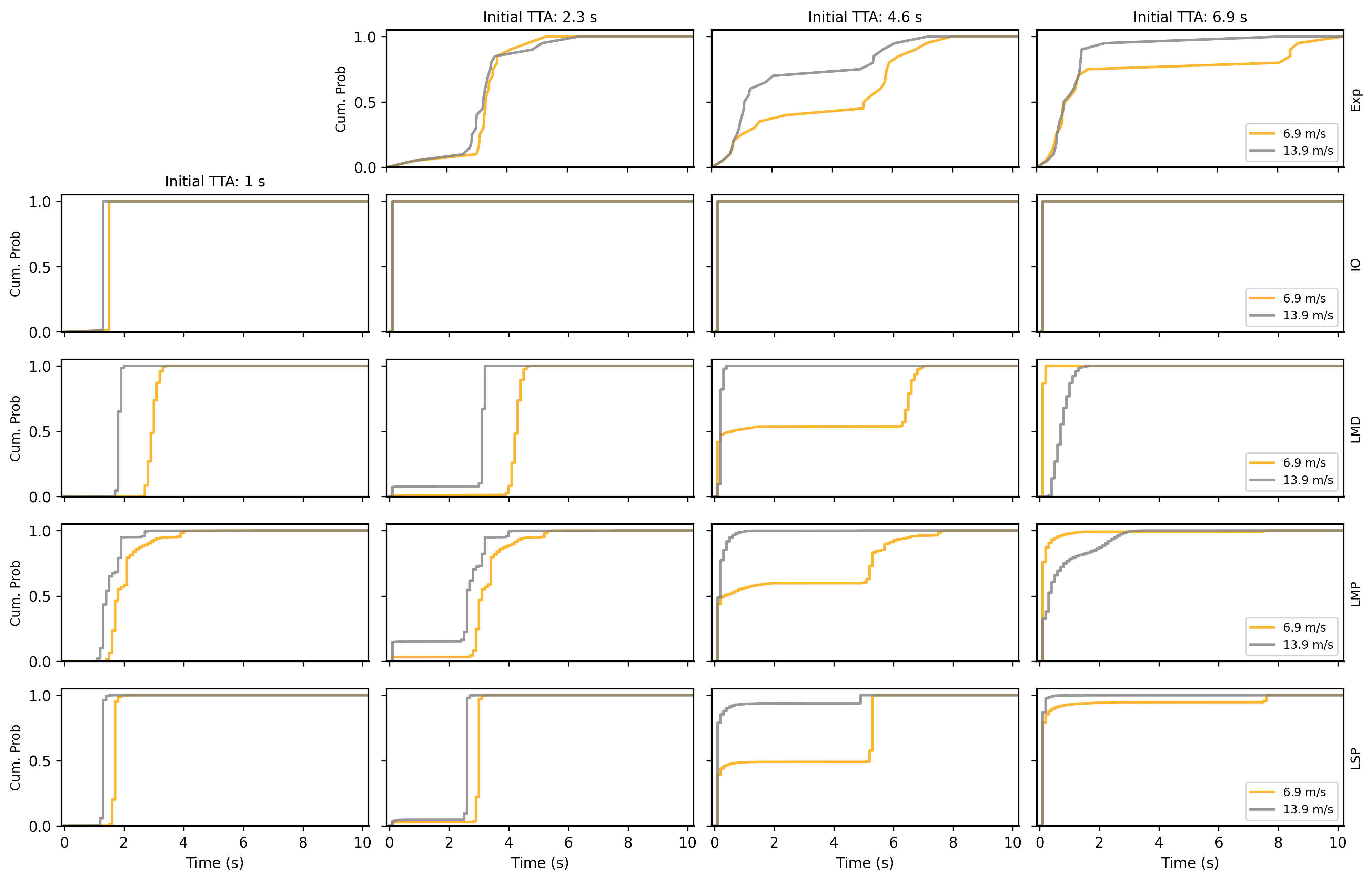}
      \vspace{-0.4cm}
      \caption{Cumulative probability for CIT. The four columns show different initial TTA conditions. See the main text for explanations of the model abbreviations. (Note: CDF curves are extended to the right after after the cumulative probability reaches $y=1$.)}
      \vspace{-0.4cm}
      \label{fig:modelcit}
\end{figure*}

The results of the model without visual limits (IO) are shown in the second panel on the left in~\figurename~\ref{fig:modelgap} and the second row in~\figurename~\ref{fig:modelcit}. In the model without visual limits, the agent had full information about the environment, and crossed the road without collision in the shortest time. Therefore, as shown in the second row of~\figurename~\ref{fig:modelcit}, the agent always started to cross at the first time step when it was safe. The second panel in~\figurename~\ref{fig:modelgap} shows that the agent always accepted the gap in $2.3~s$, $4.6~s$, and $6.9~s$ initial TTA conditions.

\begin{figure}[!b]
      \vspace{-0.2cm}  
      \centering
      \includegraphics[scale=0.35]{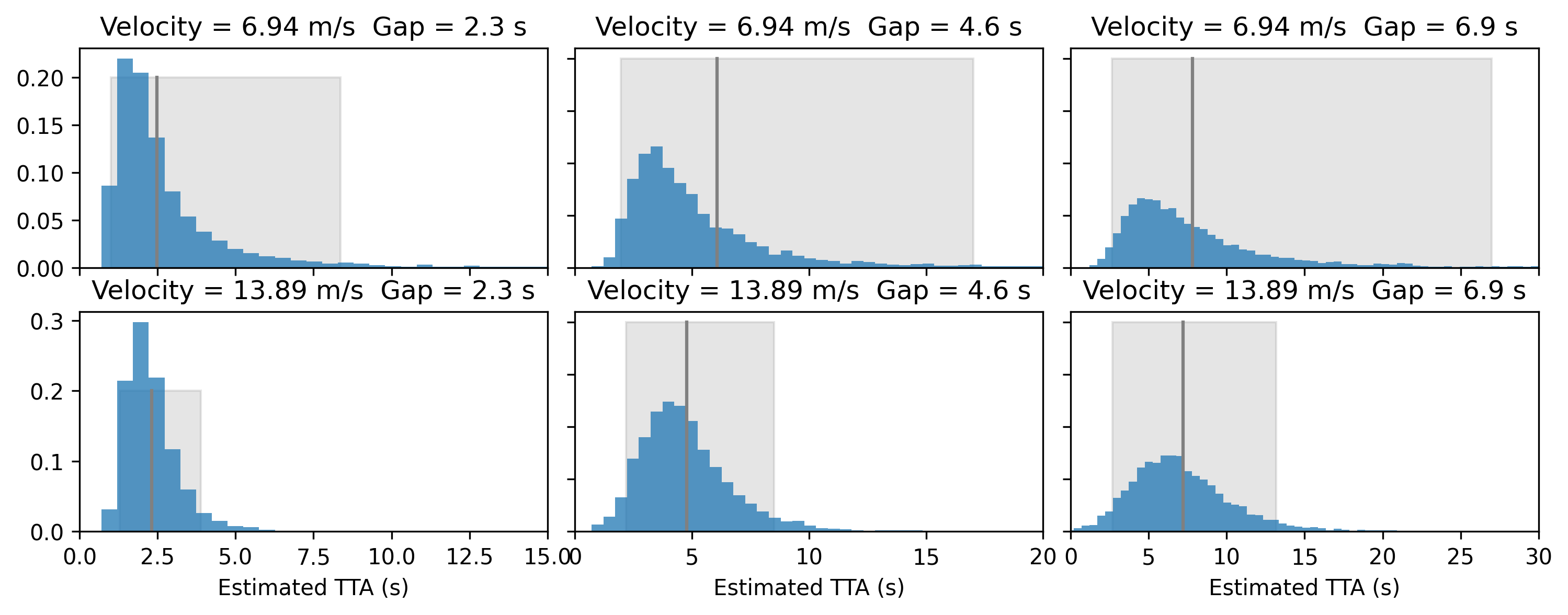}
      \vspace{-0.2cm}
      \caption{Distributions of initial estimated TTA from the output of the visual perception model, across the different scenarios in the experiment.}
      % \vspace{-0.6cm}
      \label{fig:ttadist} 
\end{figure}

The three graphs on the right in~\figurename~\ref{fig:modelgap}, and the last three rows in~\figurename~\ref{fig:modelcit} show the results of the models with visual limits. In models with visual limits, the agent received the processed information of the Kalman filter instead of the exact information about the environment. 

As shown in~\figurename~\ref{fig:modelgap}, the models with visual limits captured the pattern of the gap acceptance rate observed in the experiment. The TTA-dependent gap acceptance rate, i.e., the gap acceptance rate increased with the initial TTA, was predicted by all three model types. Unexpectedly, the difference within the same TTA group was also captured by the model. The agent was more likely to accept the gap when the vehicle speed was higher in the given initial TTA. This pattern was shown by all model types. Overall, the model has a slightly greater tendency to accept gaps than the human participants. This may be what is causing the model to show a speed dependency at $2.3~s$ TTA (because the model sometimes crosses there, whereas the humans almost never did), but not at $6.9~s$ TTA (because the model has already reached full gap acceptance at this TTA, which the humans had not). 

\subsubsection{Variability within and between individuals}
From~\figurename~\ref{fig:modelcit}, there was some spread in CIT also for model LMD – i.e., the model showed some within-individual variability – due to the trial-to-trial variability in visual noise. That model LMP, with different $\sigma_v$ for different participants, came closer to capture the human variability, suggesting that some of the variability in the human data is from between-individual differences. 

To test this, we did the Akaike Information Criterion (AIC) analysis of LMD, LMP, and LSP. AIC, which considers both the log-likelihood and the cost of more parameters, allows us to compare the performance of different models, and the preferred model is the one with the minimum AIC value~\cite{34murtaugh2014defense}. The AIC values are $113$, $57$, and $79$ for the LMD, LMP and LSP respectively. Therefore, LMP, the model with different individual $\sigma_v$, outperformed other model variants. This suggested that some of the variability in human CIT is due to between-individual variability in sensory noise.

\subsubsection{Speed-dependent gap acceptance}
To investigate the reason why the model showed speed-dependence in its gap acceptance rate, we calculated the estimated TTA through the velocity and position estimated by Kalman filter at the first time step.~\figurename~\ref{fig:ttadist} shows the distribution, the mean value, which is shown in the grey vertical line, and the $5^{th}$ and $95^{th}$ percentiles of the estimated TTA, which is shown in the shading area. As shown in~\figurename~\ref{fig:ttadist}, for the same time gap, the distribution for the estimated TTA was more dispersed at lower speed conditions. In other words, at low speeds, there was greater uncertainty about the estimated TTA, which could be the reason why it is better, from a reward maximization perspective, to be more careful about crossing in these situations. This perspective aligns with the findings of Chen et al.~\cite{36chen2021apparently}, who showed that apparently biased behavior in a more abstract choice task might also be explained as a consequence of optimal sequential decisions under uncertainty. Interestingly, Tian et al.~\cite{5tian2022explaining} showed that in the road-crossing context, humans may in practice be achieving this strategy by making use of relatively simple visual cues.
\section{Discussions and conclusion}
We developed a model of human pedestrian crossing decisions based on computational rationality, using deep RL to adopt optimal behavior policies given human-like constraints. We show that when we constrain the agent by a simple model of human visual perception, it reproduces human gap acceptance behavior qualitatively. Furthermore, the model also predicts the speed-dependencies that are typically observed in human gap acceptance. These have previously been considered as evidence of biases in human pedestrian decision-making, but our results demonstrate that this type of speed-dependence is a rational adaption to noisy visual perception. When comparing the results of LMP and LSP, it shows that these two approaches learned similar policies, but not identical policies. We believe that the approach of conditioning RL on constraint parameters is a promising approach for considering individual differences. As an early attempt in using computational rationality in modeling road user interaction, we see the feasibility to use computational rationality to model road user behavior. The developed model could be served as the agent model in the test environment of AVs.  

There is ample scope for further improvements to our model. For example, our agent is more likely to show faster responses than humans. One reason is that the human perceptual filtering might be slower than the Kalman filter we used, which can process the information without delay. Another reason is that we have not considered the non-decision time in the model. We can also observe more variability in human CITs than in model CITs. This could be due to both between- and within-individual variability of unmodeled sources. In addition, our reward function is simple and so far we have not tuned it. This is enough for our purposes, to show qualitative patterns of human road-crossing. However, this limitation is probably the reason why the model has a greater tendency for gap acceptance than humans. In future work, following a similar approach to what we did with $\sigma_v$ here, we could tune also for example the time-loss penalty in the reward function. Furthermore, in the current work, the agent is interacting with a constant speed approaching vehicle. However, in the real world, the pedestrian will interact with vehicles with various kinematic states, which also affect the crossing behavior. For example, \cite{29pekkanen2022variable} suggests that human pedestrians are also perceiving and interpreting vehicle deceleration. A major advantage of the approach we have taken here is that deep RL is scalable to much more complex traffic scenarios., far beyond what is possible with conventional cognitive models. We therefore conclude that computational rationality overall holds great promise for applied modeling of human-like road user interaction behavior.
\bibliographystyle{IEEEtran}
\bibliography{Yueyang_et_al_IEEE_IV_2023.bib}
\end{document}